\pdfoutput=1

\documentclass[11pt]{article}

\usepackage{color,soul}
\usepackage{amsmath} 
\usepackage{float}

\usepackage[]{ACL2023}

\usepackage{times}
\usepackage{latexsym}
\usepackage{adjustbox}
\usepackage{amsmath}
\usepackage{algorithm}
\usepackage{algpseudocode}
\usepackage{array}
\usepackage{multirow}
\usepackage{xcolor}
\usepackage{subcaption}

\usepackage[T1]{fontenc}

\usepackage[utf8]{inputenc}

\usepackage{microtype}

\usepackage{inconsolata}

\usepackage{paralist}

\newcommand{\td}{\textsuperscript{\textdagger}}

\title {Zero-shot Dependency Parsing with Unsupervised Cross-Lingual Bootstrapping}

\author{
  Lalita Lowphansirikul\textsuperscript{1},
  Attapol Rutherford\textsuperscript{2},
  Jian Gang Ngui\textsuperscript{3}, \\
  {\bfseries
    Sarana Nutanong\textsuperscript{1},
    Peerat Limkonchotiwat\textsuperscript{3}
  } \\[4pt]
  \textsuperscript{1}VISTEC, Thailand \quad
  \textsuperscript{2}Chulalongkorn University, Thailand \quad
  \textsuperscript{3}AI Singapore, Singapore \\[3pt]
  \texttt{lalita.l\_s22@vistec.ac.th} \quad
  \texttt{snutanon@vistec.ac.th} \\
  \texttt{attapol.t@chula.ac.th} \\
  \texttt{jiangangngui@aisingapore.org} \quad
  \texttt{peerat@aisingapore.org}
}

\begin{document}
\maketitle
\begin{abstract}

Pre-trained language models (PLMs) with encoder-based architectures have shown impressive capabilities in zero-shot cross-lingual transfer for various language understanding tasks.
However, applying this technique to dependency parsing remains a significant challenge due to its syntactic nature.
To boost model generalizability across linguistic typologies, we propose a cross-lingual unsupervised bootstrapping method to improve syntactic knowledge within the PLM.
We show that our method achieves a significant improvement in zero-shot parsing performance in low-resource languages.
Analysis of these bootstrapped models uncovers increased robustness in recognizing syntactic structures, evidenced by higher scores in parameter-free tree probing tests.

\end{abstract}

\section{Introduction}

Encoder-based multilingual pre-trained language models (PLMs) demonstrate promising cross-lingual transferability \cite{wu-dredze-2019-beto, pires-etal-2019-multilingual@how-multilingual-is-multilingual-bert}, enabling them to generalize knowledge across languages. This capability benefits language understanding tasks such as natural language inference and reasoning. 
Despite success in sequence-level tasks, cross-lingual transferability in syntactic tasks relies on syntactic similarity. \citet{tran-bisazza-2019-zero} shows that dependency parsing transferability relies on shared subject-verb-object word order. These findings highlight the importance of syntactic similarity in cross-lingual tasks.


Previous works have aimed to mitigate the structural similarity constraint in dependency parsing with various approaches.
\citet{kondratyuk-straka-2019-75@udify} propose joint training on multiple treebanks.
Further improvements include difficulty-based batch selection \cite{de-lhoneux-etal-2022-zero@depparsing_curriculum_learning}, meta-learning \cite{langedijk-etal-2022-meta@udify-maml}, and adapter modules contextualized by linguistic features \cite{ustun-etal-2020-udapter@UDapter, ji-etal-2023-typology@UDapter_position-guided}.
\citet{liu-etal-2020-cross-lingual-dependency@pos-guided_word-reordering} rearrange words based on part-of-speech tags to match target language structure.
\citet{arviv-etal-2023-improving@subtree-aware-word-reordering} improve performance with word permutation while preserving syntax.
Although these approaches have narrowed the cross-lingual transfer gap, they remain limited to supervised fine-tuning.
Our research question is as follows: \emph{``To what extent can unsupervised representation learning uplift multilingual PLM's zero-shot dependency parsing performance?''}

\begin{figure}[]
    \centering
    \includegraphics[scale=0.569]{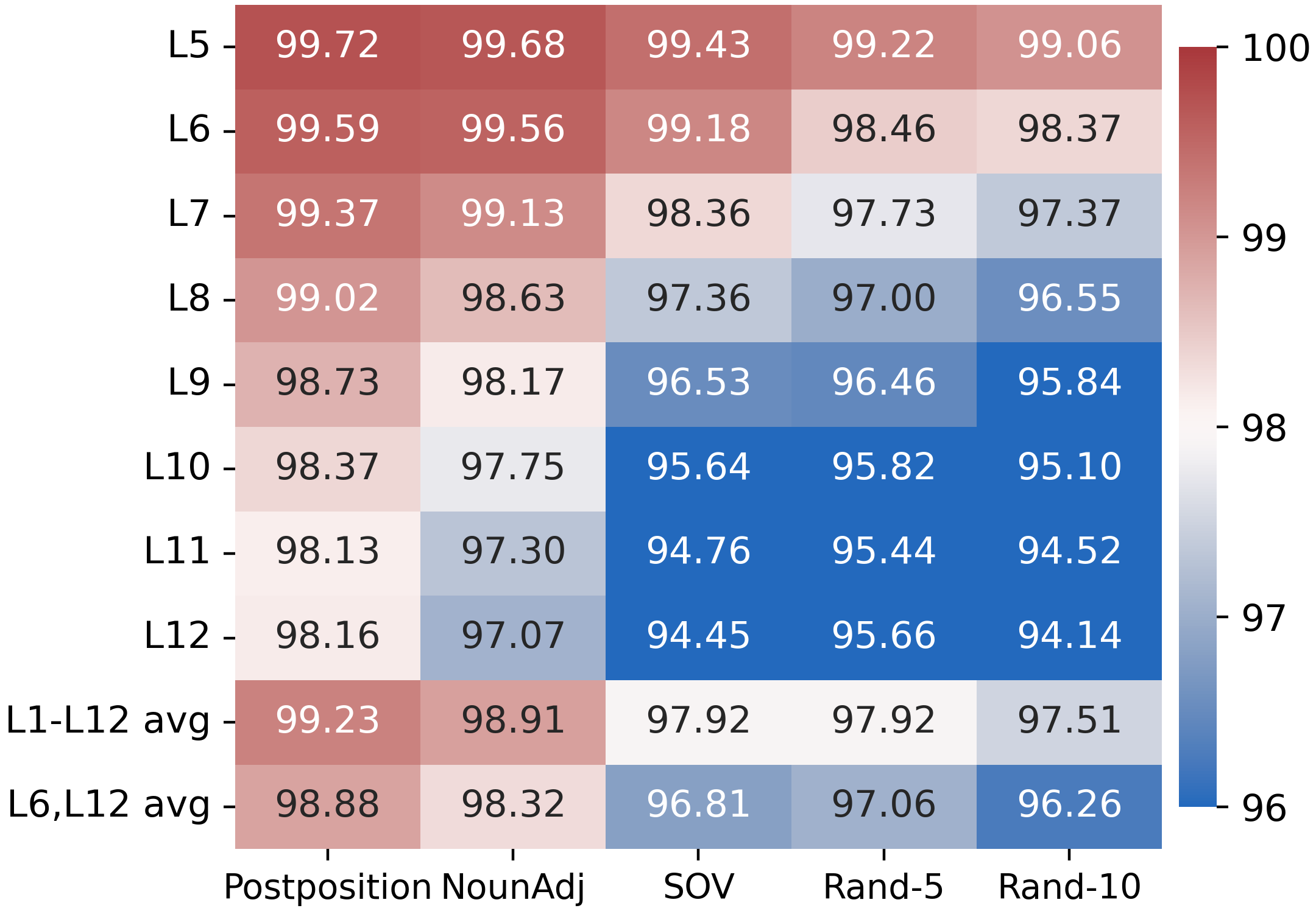}
    \caption{Layer-wise CKA similarity between original and modified English sentences obtained from pre-trained mBERT representations. A higher score means more robustness for word swapping. 
    We use the English EWT test treebank (UD v2.8) in this experiment.
    }
    \label{fig:main.layer-wise-cka_en-en_modified}
    \vspace{-5mm}
\end{figure}

We propose a PLM bootstrapping method using an unsupervised contrastive representation learning, without using multilingual datasets, to enhance representation similarity, as measured by a robust word-swap metric called centered kernel alignment (CKA)~\cite{kornblith2019similarity@CKA}, between sentences and their syntax-modified counterparts.
Our preliminary observations show pre-trained mBERT's sensitivity to word order changes (Figure~\ref{fig:main.layer-wise-cka_en-en_modified}). 
For example, the similarity between unmodified and verb-object reordered sentences decreases, similar to randomly swapped words with a 5\% distance change.
Therefore, the model training process must be invariant to various valid syntactic structures while preserving knowledge already learned.

%
%
%
We distinguish our work from previous studies by aiming to improve cross-lingual transferability through three components. 
First, we implement syntax-aware sentence augmentation by rotating subtrees, given our findings that a multilingual PLM invariant to syntactic structures can improve cross-lingual transfer for tasks requiring syntactic knowledge.
%
Second, we choose masked language modeling (MLM) as an auxiliary loss to preserve the lexicon and contextual understanding already learned in the PLM.
Finally, we apply layer combination during model bootstrapping to aggregate syntactic knowledge from other encoder layers.

%
%
We bootstrap multilingual PLMs using the components individually and in combination, then perform supervised fine-tuning with weights initialized from our bootstrapped models.
Our experimental results show improved average unlabeled and labeled attachment scores (UAS and LAS) for multilingual PLMs, especially on languages in the pre-training data. 
In addition, the experiment on probing test languages reveals higher UAS and undirected UAS (UUAS) for bootstrapped models compared to vanilla pre-trained models.


Our contributions are as follows: 
\begin{inparaenum}[(I)]
\item We present a contrastive learning-based method to improve cross-lingual transferability, incorporating syntax-aware sentence augmentation, contextual knowledge preservation, and multi-layer representation aggregation.
\item We demonstrate zero-shot downstream performance on the adapter-based parser initialized with bootstrapped models.
\item We demonstrate robust syntax changes in bootstrapped models, improving unsupervised parsing with parameter-free tree probing.
\end{inparaenum}

\section{Proposed Method}
\label{methods}
We design a bootstrapping method that enhances cross-lingual dependency parsing performance through unsupervised contrastive learning using only monolingual English data. Unlike typical contrastive setups, our approach generates syntax-aware positive samples—sentences with altered word order that retain the original syntactic meaning. This avoids the need for multilingual data, which is often unavailable for low-resource languages.
Our framework consists of three components: Syntax-aware Sentence Augmentation (Section~\ref{subsec:augmentation}, Contrastive Learning with MLM Auxiliary Loss (Section~\ref{subsec:mlm}, and Multi-layer Representation Aggregation (Section~\ref{subsec:multi_layer}).


\subsection{Syntax-aware Sentence Augmentation} \label{subsec:augmentation}
\begin{figure}[htbp]
    \centering
    \includegraphics[width=0.4285\textwidth]{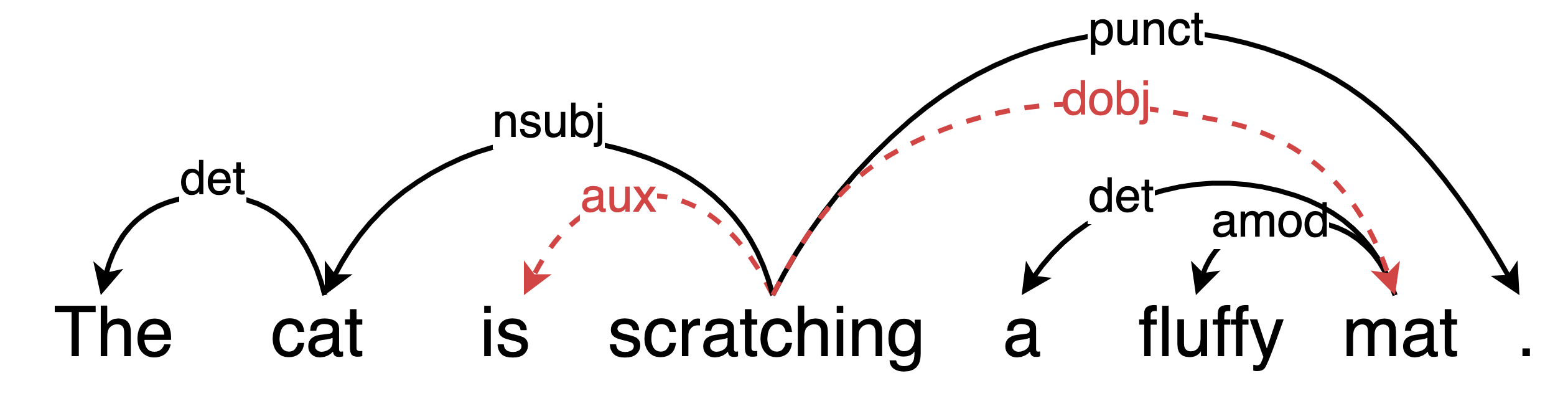}
    \includegraphics[width=0.4285\textwidth]{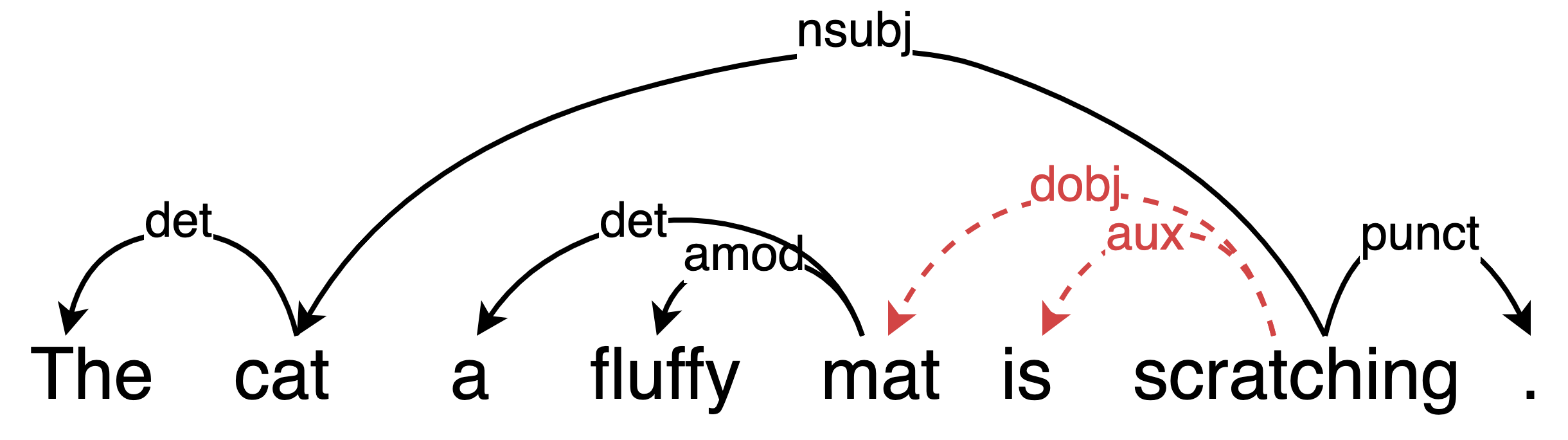}
    \caption{Original sentence (above) and OV reordered sentence (below).
Dashed arcs indicate relations related to OV reordering, while solid arcs indicate relations unrelated to OV reordering.
}
    \label{fig:example_sentence_parsed-tree.syntax-aware-augmentation.sov}
    
\end{figure}

As shown in many contrastive learning works~\cite{gao-etal-2021-simcse@simcse,wang-etal-2022-english@msimcse}, strong and robust positive samples can yield significant improvement in downstream tasks.
To create positive samples for dependency parsing, we need an augmentation scheme that significantly changes the word order while retaining the overall intended meaning of the sentence.
Therefore, we create positive samples by implementing English subtree rotation based on three word ordering features\footnote{83A: Order of Object and Verb, 85: Order of Adposition and Noun Phrase, 87A: Order of Adjective and Noun}.
%
For example, applying Object-Verb (OV) augmentation to ``The cat \emph{is scratching} a fluffy mat.'' returns ``The cat a fluffy mat \emph{is scratching}.'' (Figure \ref{fig:example_sentence_parsed-tree.syntax-aware-augmentation.sov}).
Detailed implementation is provided in the Appendix \ref{appendix:syntax_aware_sentence_augmentation}. 


%
We apply each augmentation function to original sentences, creating a one-to-one mapping. 
The resulting augmented sentences serve as positive sample candidates $x^+$ for the contrastive learning loss.
Note that we also investigate various alternatives for augmentations in Appendix~\ref{appendix:additional_exp_on_augmentation_fn_and_mlm_weight}.

\subsection{Contrastive Learning with MLM Auxiliary Loss}\label{subsec:mlm}
For robust cross-lingual representations, we employ in-batch negative contrastive learning~\cite{gao-etal-2021-simcse@simcse}.
The contrastive loss for the \textit{i}-th sentence pair in a mini-batch of $N$ pairs is given as:
\begin{equation}\label{eq:contrastive_learning}
L^{CL}_i = -\textrm{log}\frac{e^{\textrm{sim}(h^{l}_{i}, h^{+}_{i, l})/\tau}}{ \sum_{j=1}^{N} e^{\textrm{sim}(h^{l}_{i}, h^{+}_{i, l})/\tau}}    
\end{equation}
where $h^{+}_{i, l}$ denotes the encoder's sentence embedding of sentence $x^{+}_i$ from layer $l$, and $\textrm{sim}(\cdot, \cdot)$ is cosine similarity.
We use the original sentences $x_i$ as the anchor samples, augmented samples as the positive samples (Section~\ref{subsec:augmentation}), and the rest of the in-batch samples are the negative samples.

\noindent
\textbf{MLM Auxiliary Loss.}
We also apply MLM prediction \citep{devlin-etal-2019-bert} as an auxiliary loss with a scalar weight $\lambda$ to the contrastive loss to learn the grammatical understanding from the original sentence.

\begin{equation}\label{eq:contrastive_learning_with_mlm}
L_i = L^{CL}_{i} + \lambda \ L_{\textrm{mlm}}
\end{equation}

\subsection{Multi-layer Representation Aggregation} \label{subsec:multi_layer}
Based on ~\citet{hosseini-etal-2023-bert@multi-layer-simcse,oh-etal-2022-dont@dont-judge-model-by-last-layer}'s improving sentence embedding quality through layer combination, we also aggregate multi-layer representations by averaging mean-pooled representations across selected layers, rather than using only the last layer.
This results in a more robust overall representation without any additional cost for the training. 


\begin{table*}[htbp]
\centering
\scalebox{0.7}{

\begin{tabular}{|l|c| c| c|c|c|c|}
\hline
\textbf{Model} & \textbf{All} \small{(n=28)}  & \multicolumn{1}{c|}{\textbf{Unseen / }} & \multicolumn{4}{c|}{\textbf{Seen from mBERT} \small{(grouped by pre-training data size in MB)}} \\ \cline{4-7} 
 \ & Avg & Avg  & \textbf{(0, 50]}  & \textbf{(50, 100]}  & \textbf{(200, )} &  Avg   \\ \hline 

\hline

UDapter (Reproduced)            & 51.29/36.65  &  42.01/27.52  &  67.49/51.09  &  65.73/49.75  &  75.99/62.63  &  70.88/55.92 \\





UDapter-B-mBERT L12  & -0.82\td/-0.48\td         &  -0.79\td/-0.33     &  -2.07\td/-1.52        &  -0.89\td /-1.83\td  &  +0.05/+0.21     &  -0.86\td/-0.82\td \\
\ with MLM                     & +0.16/\textbf{+0.28}\td   &  -0.31\td/-0.18     &  +2.75/+3.49        &  +0.69/+0.35     &  +0.22/-0.00\td  &  +1.16/+1.24    \\
\ with SynAug OV                  & -0.30\td/+0.16            &  -1.05\td/-0.38     &  +3.70\td/+3.63\td     &  -1.41\td/-1.59     &  +0.78/+0.98 &  \textbf{+1.27}/\textbf{+1.29}\td \\
\ with SynAug OV + MLM            & \textbf{+0.16}\td /+0.25\td   &  -0.03/\textbf{+0.06}\td  &  +2.63\td /+2.15\td     &  -1.44/-0.94\td  &  +0.01/+0.31     &  +0.56/+0.65\td \\
\hline
UDapter-B-mBERT L12+L6          & +0.08/+0.04               &  +0.01\td /+0.16\td           &  -0.04\td /-1.32\td     &  +0.06/-0.15\td  &  +0.51/+0.60            &  +0.23\td /-0.21\td \\
\ with MLM                     & \textbf{+0.20}/\textbf{+0.48} &  -0.06/\textbf{+0.23} &  +1.64/+2.51\td     &  -0.72\td/-0.80     &  +0.84\td /+0.80               &  +0.76/+1.01    \\
\ with SynAug OV                 & +0.04/+0.20               &  -0.20\td/0.00               &  +1.38\td/+1.96\td     &  -0.78/-0.66     &  +0.56/+0.27     &  +0.53/+0.63    \\
\ with SynAug OV + MLM            & +0.09/+0.15               &  -0.36\td/-0.27              &  +3.05\td/+3.45\td     &  -0.82\td/-1.22     &  +0.50/+0.38     &  \textbf{+1.06}/\textbf{+1.05}    \\

\hline
\end{tabular}}
\vspace{-2mm}

\caption{Zero-shot evaluation results on 28 target languages grouped by inclusion in mBERT pre-training data and the amount of pre-training data (low-resource, middle-resource, and high-resource languages).
The scores reported are the average UAS/LAS change compared to UDapter. 
UDapter-B-mBERT denotes bootstrapped mBERT with two settings: L12 (last layer only) and L12+L6 (last and sixth layers).
%
\textdagger\ indicates a statistically significant result from the McNemar's test ($p < 0.05$).
}
\label{tab:1-main_result}
\vspace{-2mm}

\end{table*}

\section{Experimental Setup}
\vspace{-1mm}

We conduct our experiment using multilingual encoder models like mBERT and XLM-R by using our four components: syntax-aware Object-Verb augmentation, contrastive learning, MLM auxiliary loss weight, and aggregation of representations from the last and sixth layers.
Each model combination was hyperparameter-tuned for optimal similarity between original and syntax-aware sentences using English EWT treebank data \cite{silveira14gold@ud_en_ewt}. 
We also conducted with other multilingual encoders: XLM-100 \cite{@xlm100}, XLMR \cite{conneau-etal-2020-unsupervised@xlmr}, DistilBERT \cite{DBLP:journals/corr/abs-1910-01108@distilmbert}, and MiniLM \cite{wang2020minilm}. 
For boostrapping, a UDapter-based parser was fine-tuned using vanilla and bootstrapped models across 13 high-resource treebanks, utilizing trainable adapters contextualized by language typological features \cite{malaviya17emnlp@lang2vec}, and freezing model weights following \citet{ustun-etal-2020-udapter@UDapter}. 
Performance was evaluated in 28 low-resource languages using average UAS and LAS, supplemented by unsupervised tree parsing with a parameter-free tree probing method \cite{wu-etal-2020-perturbed@perturbed-masking} to assess syntactic structure capture.
We describe more details about the experimental setup in Appendix~\ref{appendix:setup}.

\section{Results and Discussion}
\vspace{-1mm}

\begin{figure*}[t]
    \centering
    \vspace{-3mm}
    \includegraphics[width=1.0\textwidth]{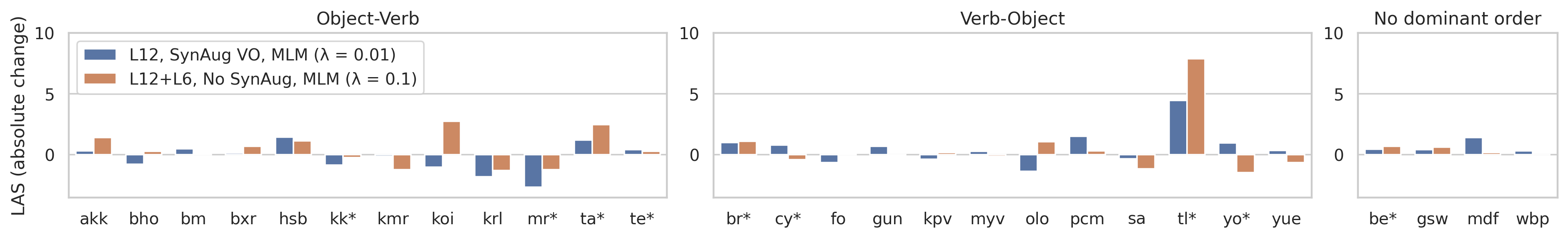}
    \vspace{-8mm}
    \caption{Zero-shot LAS parsing performance grouped by the target languages' canonical verb and object ordering.
    Results are from two UDapter models: Bootstrapped-mBERT L12 with MLM and VO augmentation, and Bootstrapped-mBERT L12+L6 with MLM loss. * denotes languages included in the mBERT pre-training data.
}
    \label{fig:downstream_performance.group_by_vo_ordering}
    \vspace{-3mm}
\end{figure*}

\noindent
\textbf{Bucket Results}. In this experiment, we split the performance of multilingual models by the amount of data (in MB) that appeared in the training data of a base model.
With this evaluation split, we can group low-resource (0,50), middle-resource (50,100), and high-resource (200-) separately.
Table~\ref{tab:1-main_result} shows that \emph{using bootstrapped models improves the performance of zero-shot parsing, especially for the (0,50) bucket}.
The bootstrapped model with last-layer representation (only L12) and OV augmentation yields the highest average UAS and LAS on the group with less data, obtaining an average improvement of +1.27 UAS and +1.29 LAS scores.
Notably, we found that combining multi-layer representation learning obtains almost similar improvement, but without any performance penalty.
This is because aggregating soft-representation from multiple layers can improve the lower-bound performance as discussed in Section~\ref{subsec:multi_layer}.
We confirm this by obtaining a decrease in performance in UDapter-B-mBERT L12 (using only vanilla contrastive learning without any of our proposed techniques).
This result emphasizes the importance of our proposed components to yield an improvement for low-resource languages.

\noindent
\textbf{Overall Performance Discussion}.
As expected, \emph{using bootstrapping increases the performance in the low-resource multilingual scenarios}.
From the overall performance, we found that we obtain significant improvements in the case of using L12 + SynAug OV + MLM, where we obtain +0.16 and +0.25 in UAS and LAS scores, respectively.
Moreover, as expected, using multi-layer aggregation also yields improvements in all performance.
We strongly believe that, with the grammatical changes from our augmentation schemes, this can also yield an improvement for large language models as well.
In conclusion, our work can bootstrap the performance of existing techniques using the unsupervised technique (contrastive learning) with our strong data augmentation (Section~\ref{subsec:augmentation}).
Note that we also experimented with other models (i.e., XLMR and DistilBERT), and the results are presented in Tables~\ref{tab:parsing.other-encoder.wiki} and \ref{tab:parsing.other-encoder.cc100}. 

\noindent
\textbf{Varied performance based on verb-object ordering.} 
To confirm the improvement seen in Table~\ref{tab:1-main_result}, we analyzed LAS differences from the baseline in three groups, based on the language's canonical verb-object ordering (Figure \ref{fig:downstream_performance.group_by_vo_ordering}).
%
While we observe overall improvement of our model, it should be noted that improvement was not uniform -- we found mixed results for OV languages while finding mostly positive results for VO languages as well as languages with no dominant order.
%

\noindent
\textbf{Unsupervised Tree Probing.}
%
%
To further understand the impact of our bootstrapping technique on improving robustness in downstream tasks, we conduct an experiment on tree probing using UAS and UUAS scores as the evaluation metric.
As shown in Table \ref{table:main.unsupervised_probing}, the bootstrapped mBERT achieves higher probed UAS and undirected UAS (UUAS) on test treebanks in single-layer and multi-layer settings. 
The Spearman's rank correlation between the uplift in probed UUAS and downstream LAS for bootstrapped models with MLM loss ($\lambda$ = 0.1), using a single-layer representation and multi-layer aggregation on seen languages, is 78.33 and 66.67, respectively ($p$ < 0.05).
%
%
Additionally, we observe that multilingual encoders pre-trained on Wikipedia show higher gains than those trained on CC-100 (Table \ref{table:additional_study_result_of_other_encoders}).
Notably, we found that the bootstrapped XLMR model's parsing performance did \emph{not} improve, which aligns with declines in its probed results.
%

\begin{table}[ht]
\centering
\scalebox{0.6}{
\vspace{-3mm}

\begin{tabular}{|l|c|c|c|}
\hline
\textbf{Treebank} & \textbf{mBERT} & \multicolumn{2}{|c|}{\textbf{Bootstrapped mBERT}}\\
\cline{3-4}
 (test set)  & \ & \textbf{L12} & \textbf{L12+L6}  \\
\hline

Unseen 				&  50.85/55.76  & -0.25/+0.95  & +0.12/+1.05  \\
Seen 				&  53.74/59.57  & +0.88/+1.85  & +1.08/+1.88 \\
\ (0, 50] 			&  53.76/58.45  & +0.21/+0.92  & +1.03/+1.34  \\
\ (50, 100] 		&  53.55/60.01  & +0.53/+0.83  & +0.46/+0.77 \\
\ (200, ) 			&  53.81/60.20  & +1.56/+3.05  & +1.43/+2.83  \\
All \small{(n=28)}  &  51.75/56.95  & +0.10/+1.23  & +0.42/+1.30  \\

\hline
\end{tabular}}
\vspace{-1mm}
\caption{Comparison of average UAS/UUAS obtained from parameter-free tree probing between vanilla mBERT and bootstrapped models trained with MLM ($\lambda$ = 0.1) and no sentence augmentation. The scores shown are from the layer with the highest UAS.
}
\label{table:main.unsupervised_probing}
\vspace{-5mm}
\end{table}



\begin{table}[t]

\centering
\scalebox{0.6}{
\begin{tabular}{|l|c|c|c|}
\hline 
\textbf{Encoder} &  \textbf{All} &  \textbf{Unseen}  & \textbf{Seen}  \\
\hline
\multirow{2}{*}{XLM-100}                  &   +0.35/+0.34              &     +0.17/-0.07           &      +0.78/+1.26   \\
\small{(Wiki)}                          & \small{(+0.26/+0.09)}    & \small{(+0.36/+0.11)}    &  \small{(+0.06/+0.04)}      \\
\multirow{2}{*}{Distil-mBERT}             &  +0.53 /+0.37            &    +1.40/+0.80          &        -0.30/-0.13    \\
\small{(Wiki, \emph{Distill})}          & \small{(+1.07/+1.82)}      & \small{(+0.80/+1.60)}     &  \small{(+1.65/+2.31)}    \\ 
\hline
\multirow{2}{*}{XLMR}                     &    -0.03/-0.39             &     -0.11/-0.36           &    -0.04/-0.52          \\
\small{(CC-100)}                        & \small{(-0.90/-1.48)}    & \small{(-0.27/-0.93)}    &  \small{(-2.05/-2.48)}   \\
\multirow{2}{*}{mMiniLM}                  &  +0.31/+0.33               &   +0.38/+0.43             &  +0.26/+0.03              \\
\small{(CC-100, \emph{Distill})}        &  \small{(+0.58/+0.44)}     & \small{(+0.54/+0.20)}     &  \small{(+0.66/+0.91)}    \\ 
\hline
\end{tabular}}
\vspace{-2mm}

\caption{Comparison of average UAS/LAS change and probed UAS/UUAS scores (in parentheses).
The results come from a bootstrapped model that uses the last layer representation, MLM auxiliary loss, and VO augmentation.
The results are selected based on the highest average UAS for the seen languages.
}
\label{table:additional_study_result_of_other_encoders}
\vspace{-5mm}

\end{table}

\section{Conclusion}

We improve cross-lingual transferability using syntax-aware sentence augmentation, MLM auxiliary loss, and layer-wise representation aggregation. 
Our method boosts zero-shot cross-lingual dependency parsing with mBERT by 1.27 UAS and 1.29 LAS on seen languages, and yields gains with other multilingual PLMs. 
Parameter-free tree probing reveals a positive link between parsing and downstream performance, suggesting that intrinsic properties of multilingual PLMs (including both encoder-based and decoder-based models) merit further study.
%


%
%

\section*{Limitations}

In this study, we focus on a model bootstrapping experiment conducted exclusively on a single source in one language, the UD English EWT treebank. 
Studies on data integration from more diverse text sources or non-English languages remain unexplored.
Regarding syntax-aware sentence augmentation, we limit our exploration to specific modifications.
Other categories of sentence augmentation, sample weights, or applying multiple modifications to the same sentence are left for future work.

We observe variability between mBERT and other pre-trained encoder-based multilingual models.
This variability can be attributed to factors such as the quantity of pre-training data available for each language and inherent properties within the models.
There is a clear need for a systematic and quantifiable methodology to identify combinations that enhance syntactic generalizability.
The potential effects on other NLU tasks, such as relation extraction and question answering, as well as experimentation on different target task fine-tuning approaches, are areas for future investigation.


\bibliography{custom}

@inproceedings{reimers-2019-sentence-bert,
    title = "Sentence-BERT: Sentence Embeddings using Siamese BERT-Networks",
    author = "Reimers, Nils and Gurevych, Iryna",
    booktitle = "Proceedings of the 2019 Conference on Empirical Methods in Natural Language Processing",
    month = "11",
    year = "2019",
    publisher = "Association for Computational Linguistics",
    url = "https://arxiv.org/abs/1908.10084",
}

@article{DBLP:journals/corr/abs-1910-01108@distilmbert,
  author       = {Victor Sanh and
                  Lysandre Debut and
                  Julien Chaumond and
                  Thomas Wolf},
  title        = {DistilBERT, a distilled version of {BERT:} smaller, faster, cheaper
                  and lighter},
  journal      = {CoRR},
  volume       = {abs/1910.01108},
  year         = {2019},
  url          = {http://arxiv.org/abs/1910.01108},
  eprinttype    = {arXiv},
  eprint       = {1910.01108},
  bibsource    = {dblp computer science bibliography, https://dblp.org}
}

@inproceedings{@xlm100,
 author = {Conneau, Alexis and Lample, Guillaume},
 booktitle = {Advances in Neural Information Processing Systems},
 editor = {H. Wallach and H. Larochelle and A. Beygelzimer and F. d\textquotesingle Alch\'{e}-Buc and E. Fox and R. Garnett},
 pages = {},
 publisher = {Curran Associates, Inc.},
 title = {Cross-lingual Language Model Pretraining},
 url = {https://proceedings.neurips.cc/paper_files/paper/2019/file/c04c19c2c2474dbf5f7ac4372c5b9af1-Paper.pdf},
 volume = {32},
 year = {2019}
}

@misc{wang2020minilm,
    title={MiniLM: Deep Self-Attention Distillation for Task-Agnostic Compression of Pre-Trained Transformers},
    author={Wenhui Wang and Furu Wei and Li Dong and Hangbo Bao and Nan Yang and Ming Zhou},
    year={2020},
    eprint={2002.10957},
    archivePrefix={arXiv},
    primaryClass={cs.CL}
}

@inproceedings{conneau-etal-2020-unsupervised@xlmr,
    title = "Unsupervised Cross-lingual Representation Learning at Scale",
    author = "Conneau, Alexis  and
      Khandelwal, Kartikay  and
      Goyal, Naman  and
      Chaudhary, Vishrav  and
      Wenzek, Guillaume  and
      Guzm{\'a}n, Francisco  and
      Grave, Edouard  and
      Ott, Myle  and
      Zettlemoyer, Luke  and
      Stoyanov, Veselin",
    editor = "Jurafsky, Dan  and
      Chai, Joyce  and
      Schluter, Natalie  and
      Tetreault, Joel",
    booktitle = "Proceedings of the 58th Annual Meeting of the Association for Computational Linguistics",
    month = jul,
    year = "2020",
    address = "Online",
    publisher = "Association for Computational Linguistics",
    url = "https://aclanthology.org/2020.acl-main.747",
    doi = "10.18653/v1/2020.acl-main.747",
    pages = "8440--8451",
}

@inproceedings{kondratyuk-straka-2019-75@udify,
    title = "75 Languages, 1 Model: Parsing {U}niversal {D}ependencies Universally",
    author = "Kondratyuk, Dan  and
      Straka, Milan",
    editor = "Inui, Kentaro  and
      Jiang, Jing  and
      Ng, Vincent  and
      Wan, Xiaojun",
    booktitle = "Proceedings of the 2019 Conference on Empirical Methods in Natural Language Processing and the 9th International Joint Conference on Natural Language Processing (EMNLP-IJCNLP)",
    month = nov,
    year = "2019",
    address = "Hong Kong, China",
    publisher = "Association for Computational Linguistics",
    url = "https://aclanthology.org/D19-1279",
    doi = "10.18653/v1/D19-1279",
    pages = "2779--2795",
}

@inproceedings{langedijk-etal-2022-meta@udify-maml,
    title = "Meta-Learning for Fast Cross-Lingual Adaptation in Dependency Parsing",
    author = "Langedijk, Anna  and
      Dankers, Verna  and
      Lippe, Phillip  and
      Bos, Sander  and
      Cardenas Guevara, Bryan  and
      Yannakoudakis, Helen  and
      Shutova, Ekaterina",
    editor = "Muresan, Smaranda  and
      Nakov, Preslav  and
      Villavicencio, Aline",
    booktitle = "Proceedings of the 60th Annual Meeting of the Association for Computational Linguistics (Volume 1: Long Papers)",
    month = may,
    year = "2022",
    address = "Dublin, Ireland",
    publisher = "Association for Computational Linguistics",
    url = "https://aclanthology.org/2022.acl-long.582",
    doi = "10.18653/v1/2022.acl-long.582",
    pages = "8503--8520",
}

@inproceedings{de-lhoneux-etal-2022-zero@depparsing_curriculum_learning,
    title = "Zero-Shot Dependency Parsing with Worst-Case Aware Automated Curriculum Learning",
    author = "de Lhoneux, Miryam  and
      Zhang, Sheng  and
      S{\o}gaard, Anders",
    editor = "Muresan, Smaranda  and
      Nakov, Preslav  and
      Villavicencio, Aline",
    booktitle = "Proceedings of the 60th Annual Meeting of the Association for Computational Linguistics (Volume 2: Short Papers)",
    month = may,
    year = "2022",
    address = "Dublin, Ireland",
    publisher = "Association for Computational Linguistics",
    url = "https://aclanthology.org/2022.acl-short.64",
    doi = "10.18653/v1/2022.acl-short.64",
    pages = "578--587",
}

@inproceedings{ustun-etal-2020-udapter@UDapter,
    title = "{UD}apter: Language Adaptation for Truly {U}niversal {D}ependency Parsing",
    author = {{\"U}st{\"u}n, Ahmet  and
      Bisazza, Arianna  and
      Bouma, Gosse  and
      van Noord, Gertjan},
    editor = "Webber, Bonnie  and
      Cohn, Trevor  and
      He, Yulan  and
      Liu, Yang",
    booktitle = "Proceedings of the 2020 Conference on Empirical Methods in Natural Language Processing (EMNLP)",
    month = nov,
    year = "2020",
    address = "Online",
    publisher = "Association for Computational Linguistics",
    url = "https://aclanthology.org/2020.emnlp-main.180",
    doi = "10.18653/v1/2020.emnlp-main.180",
    pages = "2302--2315",
}

@inproceedings{ji-etal-2023-typology@UDapter_position-guided,
    title = "Typology Guided Multilingual Position Representations: Case on Dependency Parsing",
    author = "Ji, Tao  and
      Wu, Yuanbin  and
      Wang, Xiaoling",
    editor = "Rogers, Anna  and
      Boyd-Graber, Jordan  and
      Okazaki, Naoaki",
    booktitle = "Findings of the Association for Computational Linguistics: ACL 2023",
    month = jul,
    year = "2023",
    address = "Toronto, Canada",
    publisher = "Association for Computational Linguistics",
    url = "https://aclanthology.org/2023.findings-acl.854",
    doi = "10.18653/v1/2023.findings-acl.854",
    pages = "13524--13541",
}

@inproceedings{wu-etal-2020-perturbed@perturbed-masking,
    title = "Perturbed Masking: Parameter-free Probing for Analyzing and Interpreting {BERT}",
    author = "Wu, Zhiyong  and
      Chen, Yun  and
      Kao, Ben  and
      Liu, Qun",
    editor = "Jurafsky, Dan  and
      Chai, Joyce  and
      Schluter, Natalie  and
      Tetreault, Joel",
    booktitle = "Proceedings of the 58th Annual Meeting of the Association for Computational Linguistics",
    month = jul,
    year = "2020",
    address = "Online",
    publisher = "Association for Computational Linguistics",
    url = "https://aclanthology.org/2020.acl-main.383",
    doi = "10.18653/v1/2020.acl-main.383",
    pages = "4166--4176",
}

@inproceedings{wu-dredze-2019-beto,
    title = "Beto, Bentz, Becas: The Surprising Cross-Lingual Effectiveness of {BERT}",
    author = "Wu, Shijie  and
      Dredze, Mark",
    editor = "Inui, Kentaro  and
      Jiang, Jing  and
      Ng, Vincent  and
      Wan, Xiaojun",
    booktitle = "Proceedings of the 2019 Conference on Empirical Methods in Natural Language Processing and the 9th International Joint Conference on Natural Language Processing (EMNLP-IJCNLP)",
    month = nov,
    year = "2019",
    address = "Hong Kong, China",
    publisher = "Association for Computational Linguistics",
    url = "https://aclanthology.org/D19-1077",
    doi = "10.18653/v1/D19-1077",
    pages = "833--844",
}

@inproceedings{liu-etal-2020-cross-lingual-dependency@pos-guided_word-reordering,
    title = "Cross-Lingual Dependency Parsing by {POS}-Guided Word Reordering",
    author = "Liu, Lu  and
      Zhou, Yi  and
      Xu, Jianhan  and
      Zheng, Xiaoqing  and
      Chang, Kai-Wei  and
      Huang, Xuanjing",
    editor = "Cohn, Trevor  and
      He, Yulan  and
      Liu, Yang",
    booktitle = "Findings of the Association for Computational Linguistics: EMNLP 2020",
    month = nov,
    year = "2020",
    address = "Online",
    publisher = "Association for Computational Linguistics",
    url = "https://aclanthology.org/2020.findings-emnlp.265",
    doi = "10.18653/v1/2020.findings-emnlp.265",
    pages = "2938--2948",
}

@inproceedings{tran-bisazza-2019-zero,
    title = "Zero-shot Dependency Parsing with Pre-trained Multilingual Sentence Representations",
    author = "Tran, Ke  and
      Bisazza, Arianna",
    editor = "Cherry, Colin  and
      Durrett, Greg  and
      Foster, George  and
      Haffari, Reza  and
      Khadivi, Shahram  and
      Peng, Nanyun  and
      Ren, Xiang  and
      Swayamdipta, Swabha",
    booktitle = "Proceedings of the 2nd Workshop on Deep Learning Approaches for Low-Resource NLP (DeepLo 2019)",
    month = nov,
    year = "2019",
    address = "Hong Kong, China",
    publisher = "Association for Computational Linguistics",
    url = "https://aclanthology.org/D19-6132",
    doi = "10.18653/v1/D19-6132",
    pages = "281--288",
}

@inproceedings{pires-etal-2019-multilingual@how-multilingual-is-multilingual-bert,
    title = "How Multilingual is Multilingual {BERT}?",
    author = "Pires, Telmo  and
      Schlinger, Eva  and
      Garrette, Dan",
    editor = "Korhonen, Anna  and
      Traum, David  and
      M{\`a}rquez, Llu{\'\i}s",
    booktitle = "Proceedings of the 57th Annual Meeting of the Association for Computational Linguistics",
    month = jul,
    year = "2019",
    address = "Florence, Italy",
    publisher = "Association for Computational Linguistics",
    url = "https://aclanthology.org/P19-1493",
    doi = "10.18653/v1/P19-1493",
    pages = "4996--5001",
}

@inproceedings{wang-etal-2022-english@msimcse,
    title = "{E}nglish Contrastive Learning Can Learn Universal Cross-lingual Sentence Embeddings",
    author = "Wang, Yaushian  and
      Wu, Ashley  and
      Neubig, Graham",
    editor = "Goldberg, Yoav  and
      Kozareva, Zornitsa  and
      Zhang, Yue",
    booktitle = "Proceedings of the 2022 Conference on Empirical Methods in Natural Language Processing",
    month = dec,
    year = "2022",
    address = "Abu Dhabi, United Arab Emirates",
    publisher = "Association for Computational Linguistics",
    url = "https://aclanthology.org/2022.emnlp-main.621",
    doi = "10.18653/v1/2022.emnlp-main.621",
    pages = "9122--9133",
}

@inproceedings{gao-etal-2021-simcse@simcse,
    title = "{S}im{CSE}: Simple Contrastive Learning of Sentence Embeddings",
    author = "Gao, Tianyu  and
      Yao, Xingcheng  and
      Chen, Danqi",
    editor = "Moens, Marie-Francine  and
      Huang, Xuanjing  and
      Specia, Lucia  and
      Yih, Scott Wen-tau",
    booktitle = "Proceedings of the 2021 Conference on Empirical Methods in Natural Language Processing",
    month = nov,
    year = "2021",
    address = "Online and Punta Cana, Dominican Republic",
    publisher = "Association for Computational Linguistics",
    url = "https://aclanthology.org/2021.emnlp-main.552",
    doi = "10.18653/v1/2021.emnlp-main.552",
    pages = "6894--6910",
}

@inproceedings{arviv-etal-2023-improving@subtree-aware-word-reordering,
    title = "Improving Cross-lingual Transfer through Subtree-aware Word Reordering",
    author = "Arviv, Ofir  and
      Nikolaev, Dmitry  and
      Karidi, Taelin  and
      Abend, Omri",
    editor = "Bouamor, Houda  and
      Pino, Juan  and
      Bali, Kalika",
    booktitle = "Findings of the Association for Computational Linguistics: EMNLP 2023",
    month = dec,
    year = "2023",
    address = "Singapore",
    publisher = "Association for Computational Linguistics",
    url = "https://aclanthology.org/2023.findings-emnlp.52",
    doi = "10.18653/v1/2023.findings-emnlp.52",
    pages = "718--736",
}

@inproceedings{kornblith2019similarity@CKA,
  title={Similarity of neural network representations revisited},
  author={Kornblith, Simon and Norouzi, Mohammad and Lee, Honglak and Hinton, Geoffrey},
  booktitle={International conference on machine learning},
  pages={3519--3529},
  year={2019},
  organization={PMLR}
}

@inproceedings{hosseini-etal-2023-bert@multi-layer-simcse,
    title = "{BERT} Has More to Offer: {BERT} Layers Combination Yields Better Sentence Embeddings",
    author = "Hosseini, MohammadSaleh  and
      Munia, Munawara  and
      Khan, Latifur",
    editor = "Bouamor, Houda  and
      Pino, Juan  and
      Bali, Kalika",
    booktitle = "Findings of the Association for Computational Linguistics: EMNLP 2023",
    month = dec,
    year = "2023",
    address = "Singapore",
    publisher = "Association for Computational Linguistics",
    url = "https://aclanthology.org/2023.findings-emnlp.1030",
    doi = "10.18653/v1/2023.findings-emnlp.1030",
    pages = "15419--15431",
}

@inproceedings{oh-etal-2022-dont@dont-judge-model-by-last-layer,
    title = "Don{'}t Judge a Language Model by Its Last Layer: Contrastive Learning with Layer-Wise Attention Pooling",
    author = "Oh, Dongsuk  and
      Kim, Yejin  and
      Lee, Hodong  and
      Huang, H. Howie  and
      Lim, Heuiseok",
    editor = "Calzolari, Nicoletta  and
      Huang, Chu-Ren  and
      Kim, Hansaem  and
      Pustejovsky, James  and
      Wanner, Leo  and
      Choi, Key-Sun  and
      Ryu, Pum-Mo  and
      Chen, Hsin-Hsi  and
      Donatelli, Lucia  and
      Ji, Heng  and
      Kurohashi, Sadao  and
      Paggio, Patrizia  and
      Xue, Nianwen  and
      Kim, Seokhwan  and
      Hahm, Younggyun  and
      He, Zhong  and
      Lee, Tony Kyungil  and
      Santus, Enrico  and
      Bond, Francis  and
      Na, Seung-Hoon",
    booktitle = "Proceedings of the 29th International Conference on Computational Linguistics",
    month = oct,
    year = "2022",
    address = "Gyeongju, Republic of Korea",
    publisher = "International Committee on Computational Linguistics",
    url = "https://aclanthology.org/2022.coling-1.405",
    pages = "4585--4592",
}

@inproceedings{devlin-etal-2019-bert,
    title = "{BERT}: Pre-training of Deep Bidirectional Transformers for Language Understanding",
    author = "Devlin, Jacob  and
      Chang, Ming-Wei  and
      Lee, Kenton  and
      Toutanova, Kristina",
    editor = "Burstein, Jill  and
      Doran, Christy  and
      Solorio, Thamar",
    booktitle = "Proceedings of the 2019 Conference of the North {A}merican Chapter of the Association for Computational Linguistics: Human Language Technologies, Volume 1 (Long and Short Papers)",
    month = jun,
    year = "2019",
    address = "Minneapolis, Minnesota",
    publisher = "Association for Computational Linguistics",
    url = "https://aclanthology.org/N19-1423",
    doi = "10.18653/v1/N19-1423",
    pages = "4171--4186",
}

@inproceedings{silveira14gold@ud_en_ewt,
  year = {2014},
  author = {Natalia Silveira and Timothy Dozat and Marie-Catherine de
	  Marneffe and Samuel Bowman and Miriam Connor and John Bauer and
	  Christopher D. Manning},
  title = {A Gold Standard Dependency Corpus for {E}nglish},
  booktitle = {Proceedings of the Ninth International Conference on Language
    Resources and Evaluation (LREC-2014)}
}

@inproceedings{malaviya17emnlp@lang2vec,
    title = {Learning Language Representations for Typology Prediction},
    author = {Malaviya, Chaitanya and Neubig, Graham and Littell, Patrick},
    booktitle = {Conference on Empirical Methods in Natural Language Processing (EMNLP)},
    address = {Copenhagen, Denmark},
    month = {September},
    year = {2017}
}
\bibliographystyle{acl_natbib}

\clearpage
\appendix
\section{Appendix}

\subsection{Evaluation Setup} \label{appendix:setup}

We bootstrap robust multilingual encoder models that were used in many previous cross-lingual dependency works, such as mBERT and XLM-R, using the three components detailed in Section \ref{methods}.
First, we selected Object-Verb (OV) augmentation for syntax-aware sentence augmentation due to its high CKA similarity sensitivity. 
Second, we conducted a grid search for the MLM auxiliary loss weight within \{0.1, 0.01, 0.001\}.
Lastly, we aggregated representations from the last and sixth layers due to a noticeable decrease in CKA similarity starting at the sixth layer of mBERT (Figure \ref{fig:main.layer-wise-cka_en-en_modified}). 
Thus, we used two settings: L12 (last layer) and L12+L6 (last and sixth layers).

For bootstrapping, we hyperparameter-tuned each combination to find the best model based on the average similarity between the original sentences and their three syntax-aware modifications in the final and middle-layer representations.
Sentences for model bootstrapping and validation were sourced from the train and test sets of the English EWT treebank (UD v2.8) \cite{silveira14gold@ud_en_ewt}.
Hyperparameter details are provided in Appendix \ref{appendix:model_training_details}. 
For generalizability evaluation, we conducted experiments using the same configuration on other multilingual encoders: XLM-100 \cite{@xlm100}, XLMR \cite{conneau-etal-2020-unsupervised@xlmr}, and the multilingual versions of DistilBERT \cite{DBLP:journals/corr/abs-1910-01108@distilmbert} and MiniLM \cite{wang2020minilm}.

Then, we fine-tuned a UDapter-based parser using vanilla and bootstrapped models across 13 high-resource treebanks.
UDapter employs trainable adapters contextualized by language typological features \cite{malaviya17emnlp@lang2vec} and freezes the model weights.
We followed the fine-tuning procedure from \citet{ustun-etal-2020-udapter@UDapter} and re-did sentence downsampling for UD Russian-SynTagRus since the downsampled 13k training sentences were unavailable.
We evaluated model performance in 28 low-resource languages using average UAS and LAS.
Additionally, we conducted an unsupervised tree parsing using the parameter-free tree probing method \cite{wu-etal-2020-perturbed@perturbed-masking}. This method decodes parsed trees from the token representation similarity matrix, evaluating how well the models capture syntactic structures.

\subsection{Syntax-aware sentence augmentation on English}
\label{appendix:syntax_aware_sentence_augmentation}

%
We rotate words based on three syntactic features of English using a parsed tree with UD syntactic relation annotation\footnote{https://universaldependencies.org/u/dep/index.html} while ensuring the swapped words remain within their respective subtree.  
%

\textbf{Order of Verb and Object} (VO): 
%
The object of the verb predicate is determined by nominal and clausal core arguments from the relations: \emph{iobj}, \emph{obj}, \emph{xcomp}, and \emph{ccomp}.
In our implementation, we also include non-core dependents from the relations: \emph{cop}, \emph{expl}, and \emph{obl}, and verb dependents (e.g., auxiliary verb, phrasal verb) from the relations: \emph{compound}, \emph{aux}, \emph{advmod}.
%

\textbf{Order of Adposition and Noun} (Postposition): The Adposition and Noun are determined by the following relations: \emph{case} (case marking).

\textbf{Order of Adjective and Noun} (Noun-Adj): The Adjective and Noun is determined by the following relations: \emph{amod} (adverbial modifier). 




\begin{figure}[htbp]

 
\begin{subfigure}{0.46\textwidth}
    \centering
    \includegraphics[width=\linewidth]{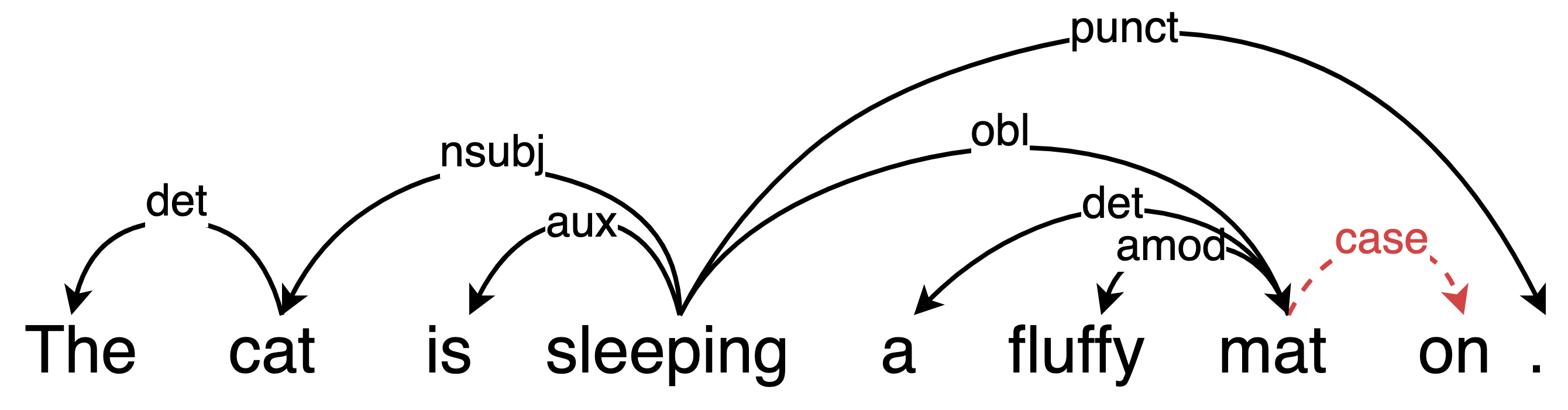}
    \caption{Noun-Adj reordered sentence}\label{fig:1a}
    \label{fig:example_sentence_parsed-tree.b}\label{fig:1a}
\end{subfigure}
\hspace{1cm} 

\begin{subfigure}{0.46\textwidth}
    \centering
    \includegraphics[width=\linewidth]{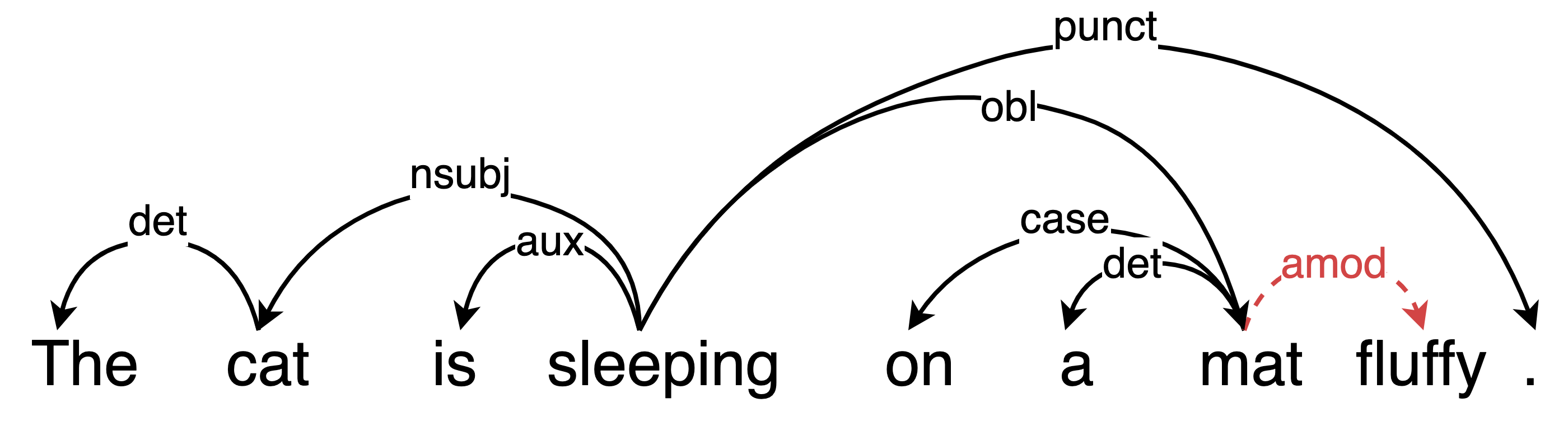}
    \caption{Adposition reordered sentence}\label{fig:1b}
    \label{fig:example_sentence_parsed-tree.c}\label{fig:1b}
\end{subfigure}
\hspace{1cm} 

\caption{Sentences with Noun-Adj reordering (a) and Adposition (b) reordering. Dashed arcs indicate relations related to reordering.}
\label{fig:example_sentence_parsed-tree.syntax-aware-augmentation}
\end{figure}

\subsection{Model training details}
\label{appendix:model_training_details}
%

We modified the sentence-transformers codebase \cite{reimers-2019-sentence-bert} and varied the hyperparameters for model bootstrapping.

\begin{compactitem}
\item Learning rate: \{ 1e-5, 1.5e-5, 3e-5 \}
\item Number of train epochs: \{ 10, 25 \}
\item Train batch size: \{ 16, 32 , 64 \}
\item Warmup ratio: 0.1
\end{compactitem}

We limit the train batch size search space to \{ 16, 32 \} for XLMR and XLM-100 due to the memory constraint.
Each run takes approximately 57 minutes for mBERT, 38 minutes for Distil-mBERT and Multilingual-MiniLM, 50 minutes for XLMR, and 97 minutes for XLM-100.
Fine-tuning the UDapter-based parser takes approximately 38 hours on mBERT, 32 on Distil-mBERT, 31 on Multilingual-MiniLM, 37 on XLMR, and 38 on XLM-100.
All experiments were conducted using NVIDIA V100 hardware except for XLM100, which ran on NVIDIA A100.

\subsection{Additional experiments on the choices of augmentation function and MLM weight}
\label{appendix:additional_exp_on_augmentation_fn_and_mlm_weight}

We report the parsing performance of bootstrapped mBERT with last-layer representation under different types of sentence augmentation in Table \ref{table:ablation_study_choices_of_augmenation_function_used} and the full results of bootstrapped mBERT for each value of the MLM loss weight in Table \ref{table:parsing.mbert.all_combinations}. Evaluation for Assyrian (aii) and Amharic (am) languages is excluded due to most tokens being unknown to mBERT's WordPiece tokenizer. 

\begin{table}[htbp]
\centering
\adjustbox{width=0.48\textwidth}{

\begin{tabular}{|l|c|c|c|}
\hline 
&  \textbf{All} &  \textbf{Unseen} & \textbf{Seen}  \\ \hline

Rand-5  &  -0.20/-0.17   	& -0.69/-0.55  		&  +0.83/+0.65               \\
Post    &  -0.20/+0.01 		& -0.35/-0.06   		& +0.13/+0.15 				 \\
OV      &  -0.30/+0.16        & -1.05/-0.38  		& \textbf{+1.27}/\textbf{+1.29}  \\
Adj     &  -0.18/-0.17          & -0.21/-0.08  	        & -0.11/-0.35 	                 \\
OV, Adj &  -0.21/-0.08          &  -0.36/-0.19 	        & +0.10/+0.16	                 \\
\hline

\end{tabular}}
\caption{Comparison in average UAS/LAS change with different augmentations between baseline and Bootstraped-mBERT Single-Layer without MLM.
%
}
\label{table:ablation_study_choices_of_augmenation_function_used}
\end{table}

\begin{table*}[]
\centering
\scalebox{0.6}{

\begin{tabular}{|l|c| c| c|c|c|c|}
\hline
\textbf{Model} & \textbf{All} \small{(n=28)}  & \multicolumn{1}{c|}{\textbf{Unseen / }} & \multicolumn{4}{c|}{\textbf{Seen from mBERT} \small{(grouped by pre-training data size in MB)}} \\ \cline{4-7} 
 \ & Avg & Avg  & \textbf{(0, 50]}  & \textbf{(50, 100]}  & \textbf{(200, )} &  Avg   \\ \hline 

\hline
UDapter (Reproduced)           & 51.29/36.65  &  42.01/27.52  &  67.49/51.09  &  65.73/49.75  &  75.99/62.63  &  70.88/55.92 \\
\hline

UDapter-B-mBERT L12 & -0.82\td/-0.48\td  &  -0.79\td/-0.33     &  -2.07\td/-1.52     &  -0.89\td/-1.83\td  &  +0.05/+0.21     &  -0.86\td/-0.82\td \\

\ MLM \small{($\lambda=0.1$)}         & +0.16/+0.28\td  &  -0.31\td/-0.18     &  +2.75/+3.49     &  +0.69/+0.35     &  +0.22/-0.00\td  &  +1.16/+1.24    \\
\ MLM \small{($\lambda=0.01$)}			& -0.38/+0.05     &  -0.81\td/-0.14     &  +1.71\td/+2.03\td  &  +0.21/-1.13     &  -0.14\td/+0.08\td  &  +0.55\td/+0.46\td \\
\ MLM \small{($\lambda=0.001$)}			& -1.07\td/-0.93\td  &  -1.50\td/-0.81\td  &  -0.84\td/-2.22\td  &  +1.64/-1.03\td  &  -0.58\td/-0.51\td  &  -0.17\td/-1.20\td \\
\hline
\ SynAug      						    & -0.30\td/+0.16     &  -1.05\td/-0.38     &  +3.70\td/+3.63\td  &  -1.41\td/-1.59     &  +0.78/+0.98     &  +1.27/+1.29\td \\
\ SynAug + MLM \small{($\lambda=0.1$)} 	  & -0.36/+0.04     &  -0.74\td/-0.38     &  +1.26/+2.62\td  &  -0.71/-0.78     &  +0.37/+0.51     &  +0.43/+0.93    \\
\ SynAug + MLM \small{($\lambda=0.01$)}     & +0.16\td/+0.25\td  &  -0.03/+0.06\td  &  +2.63\td/+2.15\td  &  -1.44/-0.94\td  &  +0.01/+0.31     &  +0.56/+0.65\td \\
\ SynAug + MLM \small{($\lambda=0.001$)}    & -0.23\td/-0.18\td  &  -0.45\td/-0.24\td  &  +0.67/+0.86\td  &  -0.23/-1.59     &  +0.10/+0.03     &  +0.21/-0.05    \\

\hline

UDapter-B-mBERT L12+L6    & +0.08/+0.04     &  +0.01\td/+0.16\td  &  -0.04\td/-1.32\td  &  +0.06/-0.15\td  &  +0.51/+0.60     &  +0.23\td/-0.21\td \\

\ MLM \small{($\lambda=0.1$)}    & +0.20/+0.48     &  -0.06/+0.23     &  +1.64/+2.51\td  &  -0.72\td/-0.80     &  +0.84\td/+0.80     &  +0.76/+1.01    \\
\ MLM \small{($\lambda=0.01$)}    & +0.07\td/+0.06     &  -0.25/0.00      &  +1.77/+0.32\td  &  -0.54\td/-0.09     &  +0.66/+0.18\td  &  +0.76\td/+0.17    \\
\ MLM \small{($\lambda=0.001$)}   & -0.30/-0.00     &  -0.33\td/-0.09     &  +0.41/+1.70     &  -2.22\td/-1.78\td  &  +0.29/-0.00     &  -0.22\td/+0.17\td \\
\hline
\ SynAug 								 & +0.04/+0.20     &  -0.20\td/0.00      &  +1.38\td/+1.96\td  &  -0.78/-0.66     &  +0.56/+0.27     &  +0.53/+0.63    \\
\ SynAug + MLM \small{($\lambda=0.1$)}	 & +0.09/+0.15     &  -0.36\td/-0.27     &  +3.05\td/+3.45\td  &  -0.82\td/-1.22     &  +0.50/+0.38     &  +1.06/+1.05    \\
\ SynAug + MLM \small{($\lambda=0.01$)}  & -1.37\td/-1.07\td  &  -1.34\td/-0.63\td  &  -2.28\td/-3.94\td  &  -2.48\td/-2.77\td  &  -0.31\td/-0.17\td  &  -1.45\td/-2.00\td \\
\ SynAug + MLM \small{($\lambda=0.001$)} & -0.77\td/-0.42\td  &  -1.05\td/-0.54\td  &  +0.17/-0.18     &  -0.36\td/-0.21     &  -0.36\td/-0.17\td  &  -0.18\td/-0.18\td \\

\hline
 
\end{tabular}}

\caption{Zero-shot evaluation result on 28 target languages grouped by inclusion in mBERT pre-training data and the amount of pre-training data.
The scores reported are the average UAS/LAS change compared to UDapter.
UDapter-B-mBERT denotes bootstrapped mBERT with two settings: L12 (last layer only) and L12+L6 (last and sixth layers).
The $\lambda$ denotes the weight assigned to the MLM auxiliary loss during the model bootstrapping phase.
\textdagger\ indicates a statistically significant result from McNemar's test ($p < 0.05$).
The pre-training data sizes (in MB) for seen languages are as follows: Breton (be), Tagalog (tl), Yoruba (yo) (0-50 MB); Marathi (mr), Welsh (cy) (50-100 MB); Belarusian (be), Kazakh (kk), Tamil (ta), Telugu (te) (>200 MB).
Data sizes are calculated from the Wikipedia dump plaintext (snapshot: 2018-02-25).
}
\label{table:parsing.mbert.all_combinations}
\end{table*}


\newpage
\subsection{Zero-shot dependency parsing performance of other multilingual encoders.}
\label{appendix:parsing.other-encoders}
We report the parsing performance of multilingual encoders pre-trained on Wikipedia data in Table \ref{tab:parsing.other-encoder.wiki} and CC-100 in Table \ref{tab:parsing.other-encoder.cc100}.

\begin{table*}[]
\centering
\scalebox{0.6}{

\begin{tabular}{|l|c| c| c|c|c|c|}
\hline
\textbf{Model} & \textbf{All} \small{(n=28)}  & \multicolumn{1}{c|}{\textbf{Unseen / }} & \multicolumn{4}{c|}{\textbf{Seen from the encoder} \small{(grouped by pre-training data size in MB)}} \\ \cline{4-7} 
 \ & Avg & Avg  & \textbf{(0, 50]}  & \textbf{(50, 100]}  & \textbf{(200, )} &  Avg   \\ \hline 

\hline
UDapter-Distil-MBert  & 45.73/30.74  &  39.31/25.29  &  50.13/31.34  &  56.24/37.45  &  67.65/52.84  &  59.27/42.25 \\
UDapter XLM-100                & 49.87/36.18  &  43.67/29.92  &  55.25/41.61  &  62.33/46.76  &  69.08/56.51  &  62.97/49.38 \\

\hline
UDapter-Distil-MBert  L6   						& +0.30\td/+0.20\td  &  +1.41\td/+0.82\td  &  -0.94\td/+0.05     &  -1.92\td/-2.51\td  &  -0.41/-0.45     &  -0.92/-0.67\td \\
\ MLM \small{($\lambda=0.01$)}					& +0.74\td/+0.49\td  &  +1.68\td/+0.62\td  &  -1.53\td/+0.81\td  &  +1.18/+0.33     &  -0.24/-0.02     &  -0.47\td/+0.38\td \\
\ SynAug                     				& +0.58\td/+0.31     &  +1.40\td/+0.64\td  &  +0.55/+0.45\td  &  -0.18\td/-1.62\td  &  +0.56\td/+0.55\td  &  +0.40\td/+0.08\td \\
\ SynAug + MLM \small{($\lambda=0.01$)}     & +0.73\td/+0.48\td  &  +1.76\td/+1.02\td  &  -1.84\td/-0.96\td  &  +0.97\td/+0.38     &  +0.28/+0.07     &  -0.43\td/-0.28\td \\

\hline
UDapter XLM-100 L16         & +0.86\td/+0.60     &  +1.23\td/+0.65\td  &  +0.04\td/+0.74\td  &  +0.80/+0.11\td  &  +0.61\td/+1.05\td  &  +0.23\td/+0.57\td \\
\ MLM \small{($\lambda=0.001$)}	              & +2.08\td/+1.66\td  &  +1.87\td/+1.11\td  &  +3.89\td/+5.33\td  &  +2.65\td/+1.91\td  &  +1.76\td/+1.84\td  &  +2.32\td/+2.70\td \\
\ SynAug                                    & -  &  -  &  -  &  -  &  - & - \\
\ SynAug + MLM \small{($\lambda=0.1$)}     & +0.54/+0.41     &  +0.40/-0.00     &  -0.24\td/+0.18\td  &  -0.17/+0.07     &  +1.65\td/+2.23\td  &  +0.52/+0.92    \\
\hline
\hline
UDapter XLM-100 L16+L8                              & -  &  -  &  -  &  -  &  - & - \\
\ MLM \small{($\lambda=0.1$)}                    & -0.30\td/-0.03\td  &  +0.09\td/+0.37     &  -0.20\td/-0.75\td  &  -0.28/-0.75\td  &  -1.29\td/-0.37\td  &  -0.79\td/-0.58\td \\
\ SynAug                                           & +0.26\td/+0.09\td  &  +0.48/+0.02\td  &  +0.16\td/+0.62\td  &  -0.65\td/-0.90\td  &  -0.88\td/+0.01\td  &  -0.39\td/+0.02\td \\
\ SynAug + MLM \small{($\lambda=0.001$)}	       & +0.85\td/+0.47     &  +0.76\td/+0.16     &  +2.88\td/+2.55\td  &  +1.46\td/+0.24\td  &  -0.24/+0.80\td  &  +1.12\td/+1.13    \\

\hline

\hline
\end{tabular}}
\caption{Zero-shot evaluation result on 28 target languages grouped by inclusion in mBERT pre-training data and the amount of pre-training data.
The scores reported are the average UAS/LAS change compared to UDapter. UDapter-B-Distil-mBERT denotes bootstrapped mBERT.
The $\lambda$ denotes the weight assigned to the MLM auxiliary loss during the model bootstrapping phase. 
The row marked with "-" indicates that the model failed to converge.
\textdagger\ indicates a statistically significant result from McNemar's test ($p < 0.05$).
}
\label{tab:parsing.other-encoder.wiki}
\end{table*}


\begin{table*}[]
\centering
\scalebox{0.6}{

\begin{tabular}{|l|c| c| c|c|c|c|}
\hline
\textbf{Model} & \textbf{All} \small{(n=28)}  & \multicolumn{1}{c|}{\textbf{Unseen / }} & \multicolumn{4}{c|}{\textbf{Seen from XLMR} \small{(grouped by pre-training data size in GB)}} \\ \cline{4-7} 
 \ & Avg & Avg  & \textbf{(0, 4]}  & \textbf{(4, 10]}  & \textbf{(10, )} &  Avg   \\ \hline 

\hline
UDapter-mMiniLM                 & 51.50/36.85  &  40.52/26.53  &  56.86/35.80  &  75.94/61.09  &  80.01/72.22  &  68.01/51.62 \\

UDapter-XLMR                   & 53.96/40.20  &  41.03/27.80  &  66.71/47.05  &  78.57/64.03  &  81.58/73.84  &  73.73/58.10 \\
 
\hline
UDapter-B-XLMR L12          & -1.23/-0.86  &  -1.67/-0.98  &  -0.64/-0.57  &  -0.40/-0.61  &  -0.62/-0.45  &  -0.55/-0.56 \\
\ MLM \small{($\lambda=0.01$)}                 & +0.14/-0.27  &  +0.10/-0.38  &  -0.36/-0.58  &  +0.29/+0.30  &  +0.29/+0.26  &  -0.01/-0.11 \\
\ SynAug                  & -0.44/-0.48  &  -0.18/-0.21  &  -0.69/-0.94  &  -0.48/-1.31  &  +0.48/+0.40  &  -0.40/-0.83 \\

\ SynAug + MLM \small{($\lambda=0.05$)}     & -0.46/-0.35  &  -0.76/-0.30  &  -0.72/-0.77  &  +0.36/-0.38  &  +0.08/+0.10  &  -0.18/-0.47 \\
\hline
UDapter-B-mMiniLM L12        & +0.34/+0.36  &  +0.24/+0.42  &  -0.42/-0.36  &  +1.03/+0.33  &  +0.22/+0.06  &  +0.23/-0.03 \\
\ MLM \small{($\lambda=0.01$)}           & -0.24/+0.16  &  -0.51/+0.15  &  -0.59/-0.28  &  +1.04/+0.70  &  +0.09/-0.29  &  +0.13/+0.07 \\
\ SynAug             & +0.16/+0.35  &  +0.13/+0.44  &  -0.87/-0.53  &  +1.35/+0.74  &  -0.10/-0.10  &  +0.08/+0.01 \\
\ SynAug + MLM \small{($\lambda=0.001$)} & +0.31/+0.33  &  +0.38/+0.43  &  -0.26/-0.15  &  +0.97/+0.35  &  +0.16/-0.17  &  +0.26/+0.03 \\

\hline
\hline
UDapter-B-XLMR L12+L6        & -1.21/-0.68  &  -1.70/-0.86  &  -0.75/-0.57  &  -0.13/-0.05  &  -0.26/+0.01  &  -0.44/-0.28 \\
\ MLM \small{($\lambda=0.05$)}                & -0.23/-0.41  &  -0.25/-0.02  &  -1.24/-1.80  &  +0.04/-0.94  &  -0.20/-0.07  &  -0.59/-1.17 \\
\ SynAug                  & -0.38/-0.38  &  -0.33/-0.24  &  -1.28/-1.26  &  -0.52/-0.47  &  +0.21/+0.10  &  -0.74/-0.73 \\

\ SynAug + MLM \small{($\lambda=0.1$)}      & +0.01/-0.36  &  -0.00/-0.27  &  -0.51/-1.04  &  +0.80/+0.02  &  +0.04/-0.02  &  +0.06/-0.47 \\
\hline
UDapter-B-MiniLM L12+L6      & -0.53/+0.02  &  -0.72/-0.05  &  -0.84/-0.01  &  +0.39/+0.18  &  -0.35/-0.31  &  -0.30/0.00 \\
\ MLM \small{($\lambda=0.05$)}          & +0.35/+0.41  &  +0.49/+0.48  &  -0.88/-0.50  &  +0.87/+0.66  &  +0.30/-0.25  &  -0.03/-0.03 \\
\ SynAug             & -0.12/+0.19  &  -0.20/+0.28  &  -1.04/-0.83  &  +0.92/+0.61  &  +0.18/+0.02  &  -0.10/-0.15 \\
\ SynAug + MLM \small{($\lambda=0.001$)} & +0.15/+0.40  &  +0.02/+0.38  &  -0.50/-0.21  &  +1.60/+1.09  &  0.00/+0.05  &  +0.36/+0.31 \\

\hline

\hline
\end{tabular}}
\caption{Zero-shot evaluation result on 28 target languages grouped by inclusion in mBERT pre-training data and the amount of pre-training data.
The scores reported are the average UAS/LAS change compared to UDapter. UDapter-B-XLMR denotes bootstrapped XLMR. UDapter-B-mMiniLM denotes bootstrapped Multilingual MiniLM. The $\lambda$ denotes the weight assigned to the MLM auxiliary loss during the model bootstrapping phase.
\textdagger\ indicates a statistically significant result from McNemar's test ($p < 0.05$).
}
\label{tab:parsing.other-encoder.cc100}
\end{table*}

\begin{table*}[htbp]
\footnotesize
    \centering
    \scalebox{0.6}{
    \begin{tabular}{|l|c|c|c|c|c|c|c|}
\hline
  &  \textbf{UDapter} & \multicolumn{6}{|c|}{\textbf{UDapter-Bootstrapped mBERT}} \\ \cline{3-8}
 &  (Baseline) & \multicolumn{3}{|c|}{\textbf{Single-Layer (L12)}} & \multicolumn{3}{|c|}{ \textbf{Multi-Layer (L12+L6)}} \\  \cline{3-8}
 &  & \textbf{MLM} & \textbf{SynAug} & \textbf{MLM} & \textbf{MLM} & \textbf{SynAug} & \textbf{MLM} \\
 &  & \ &   & \textbf{+SynAug} &   &   & \textbf{+SynAug} \\ 
\hline

akk   & 42.38/49.11 & -0.12/+0.40 & -1.04/-0.12 & +0.58/+1.15 & -0.81/-0.86 & -4.89/-6.10 & +0.06/-0.29 \\ 
be$^*$  & 34.9/47.3 & +2.58/+4.40 & -0.11/-1.62 & +1.67/+3.81 & +2.61/+4.20 & -1.44/-2.12 & +1.44/+0.63 \\ 
bho   & 40.09/48.86 & -0.06/+1.27 & -0.23/+1.82 & -0.55/+0.80 & -0.90/+0.55 & -2.44/+1.15 & -0.57/+0.57 \\ 
bm    & 43.79/50.19 & +0.25/+0.91 & -0.18/+0.81 & +0.41/+1.14 & +0.94/+1.33 & +0.47/+0.96 & +0.49/+0.56 \\ 
br$^*$  & 57.29/61.33 & -1.57/+0.17 & -4.37/-1.23 & -1.53/+0.24 & -0.78/+0.75 & -3.91/-3.27 & -1.67/+0.24 \\ 
bxr   & 50.92/56.64 & +1.96/+3.00 & +2.01/+2.80 & +1.63/+3.26 & +1.51/+2.73 & -0.30/+1.37 & +1.29/+2.78 \\ 
cy$^*$  & 44.48/53.51 & +0.08/+0.44 & -3.15/-1.88 & -0.72/-0.18 & +0.18/+0.32 & -1.89/-1.85 & -0.49/-0.55 \\ 
fo    & 62.63/63.85 & +0.15/+1.35 & -0.53/+0.67 & +0.25/+1.45 & +0.32/+1.57 & -1.63/-0.85 & +0.11/+1.27 \\ 
gsw   & 51.87/54.64 & +0.07/+1.94 & -2.08/-0.83 & -0.42/+1.52 & -0.07/+1.18 & -3.12/-2.22 & +0.21/+1.80 \\ 
gun   & 46.05/50.99 & +0.91/+1.14 & +0.83/+0.76 & +2.58/+3.03 & +0.91/+1.14 & +1.21/+0.46 & +0.23/+1.67 \\ 
hsb   & 38.05/45.25 & +0.36/+0.59 & -0.62/-0.88 & +0.51/+0.45 & +0.73/+0.49 & -2.34/-2.16 & +0.75/+0.34 \\ 
kk$^*$  & 56.24/59.92 & +1.10/+2.70 & -0.79/+0.27 & +0.10/+1.97 & +0.73/+2.18 & -1.67/-0.27 & +0.08/+1.64 \\ 
kmr   & 44.9/50.23 & +0.35/+2.04 & -0.12/+1.03 & +0.15/+1.16 & +0.57/+2.05 & +0.07/+0.13 & +0.83/+0.89 \\ 
koi   & 58.15/59.65 & -1.00/+0.75 & -1.75/+0.25 & 0.00/+2.01 & -0.75/+0.75 & -3.01/-1.75 & +1.00/+2.01 \\ 
kpv   & 53.3/58.12 & -2.18/+1.40 & -0.78/+1.55 & -1.55/+1.09 & -0.70/+1.71 & -0.93/+0.70 & -1.17/+1.55 \\ 
krl   & 46.9/53.62 & +0.71/+0.52 & -0.87/-1.68 & +0.32/-0.94 & +0.58/+0.68 & -0.87/-1.29 & -0.06/-1.42 \\ 
mdf   & 54.72/57.04 & +1.96/+3.21 & +2.14/+3.39 & +2.14/+3.03 & +4.10/+4.99 & +0.89/+2.85 & +3.21/+3.74 \\ 
mr$^*$  & 62.62/66.5 & +0.97/+1.21 & -0.24/-0.73 & -0.73/+0.24 & +0.73/+1.21 & +0.73/+0.73 & +1.21/+0.97 \\ 
myv   & 52.92/58.24 & -0.32/+0.35 & -0.27/-0.58 & -0.52/+0.72 & +0.61/+1.53 & -0.94/-0.48 & -0.18/+0.04 \\ 
olo   & 43.01/50.0 & +1.75/+1.28 & +0.20/-0.34 & +1.14/+0.81 & +1.61/+1.21 & -0.47/-1.14 & +0.94/+0.74 \\ 
pcm   & 43.55/49.44 & -0.79/+0.60 & -1.45/+0.11 & -0.16/+0.71 & -0.68/+0.03 & -1.78/-0.68 & -0.71/+0.14 \\ 
sa    & 57.89/60.39 & -1.63/-0.11 & -0.49/-1.79 & -1.95/+0.16 & -1.36/-0.27 & -0.65/-2.01 & -1.68/-0.43 \\ 
ta$^*$  & 42.28/51.73 & +2.01/+3.02 & +1.71/+2.11 & +1.66/+2.77 & +1.01/+2.46 & -0.55/+1.81 & +0.45/+2.16 \\ 
te$^*$  & 81.83/81.83 & +0.55/+2.08 & +0.69/+1.11 & +0.83/+2.08 & +1.39/+2.50 & -0.28/+0.28 & +0.69/+1.94 \\ 
tl$^*$  & 73.97/73.97 & +1.71/+2.74 & +1.71/+1.71 & +1.71/+3.08 & +2.74/+2.74 & 0.00/0.00 & +1.03/+1.03 \\ 
wbp   & 90.45/90.45 & -1.91/-1.91 & +0.64/+0.64 & -0.96/-0.96 & -0.64/-0.64 & 0.00/0.00 & -1.27/-1.27 \\ 
yo$^*$  & 30.03/40.05 & +0.49/-0.15 & +0.68/+0.08 & +0.49/+1.39 & +1.13/+0.53 & +0.68/-0.34 & +0.64/+0.08 \\ 
yue   & 45.03/51.22 & -2.66/-0.19 & -0.78/+0.06 & -1.31/+0.70 & -1.80/+0.26 & +0.50/+1.54 & -1.83/+0.32 \\ 

 
\hline
    \end{tabular}}
    \caption{Comparison of per-language probing UAS/UUAS between vanilla mBERT and bootstrapped models on test languages.
The test languages include Akkadian (akk), Bambara (bm), Belarusian (be), Bhojpuri (bho), Breton (be), Buryat (bxr), Cantonese (yue), Erzya (myv), Faroese (fo), Karelian (krl), Kazakh (kk), Komi Permyak (koi), Komi Zyrian (kpv), Kurmanji (kmr), Livvi (olo), Marathi (Mr), Mbya Guarani (gun), Moksha (mdf), Nigerian Pidgin (pcm), Sanskrit (sa), Swiss German (gsw), Tagalog (tl), Tamil (ta), Telugu (te), Upper Sorbian (hsb), Warlpiri (wbp), Welsh (cy), and Yoruba (yo).
Language codes with asterisks (*) indicate languages included in the mBERT pre-training data. 
}
    \label{table:probing_result_per_language.out-mbert}
\end{table*}

\begin{table*}[]
\footnotesize
\centering
\scalebox{0.6}{
    \begin{tabular}{|l|c|c|c|c|c|c|c|}
\hline
  &  \textbf{UDapter} & \multicolumn{6}{|c|}{\textbf{UDapter-Bootstrapped mBERT}} \\ \cline{3-8}
 &  (Baseline) & \multicolumn{3}{|c|}{\textbf{Single-Layer (L12)}} & \multicolumn{3}{|c|}{ \textbf{Multi-Layer (L12+L6)}} \\  \cline{3-8}
 &  & \textbf{MLM} & \textbf{SynAug} & \textbf{MLM} & \textbf{MLM} & \textbf{SynAug} & \textbf{MLM} \\
 &  & \ &   & \textbf{+SynAug} &   &   & \textbf{+SynAug} \\ \hline

akk & 25.76/5.56 & -0.22/+1.24\td & -9.29\td/-2.00\td & -1.94/+0.32 & -1.35/+1.40\td & -7.13\td/-1.46\td & -0.54/+1.73\td   \\
be$^*$ & 84.31/79.54 & +0.28/+0.52 & +0.52/+0.66\td & +0.11/+0.46 & +0.37/+0.70\td & +0.07/+0.48 & +0.11/+0.42   \\
bho & 51.75/36.43 & -0.10/+0.08 & -0.49/-0.31 & -0.84\td/-0.76\td & -0.14/+0.29 & -0.14/-0.20 & -0.47/-0.18   \\
bm & 27.99/7.97 & +1.24\td/-0.54\td & +0.97\td/-0.43\td & +1.06\td/+0.49\td & +1.37\td/+0.12 & +0.17/-0.72\td & +1.54\td/-0.34   \\
br$^*$ & 71.22/55.65 & -0.21/+0.28 & +1.00\td/+2.00\td & +0.92\td/+1.01\td & +0.39/+1.10\td & +1.08\td/+1.58\td & +0.94\td/+1.21\td   \\
bxr & 48.89/28.66 & -0.58/-0.63\td & -1.02\td/-1.08\td & -0.18/+0.14 & +1.02\td/+0.71\td & -0.54/-0.88\td & -0.15/-0.16   \\
cy$^*$ & 71.50/54.84 & -0.08/-0.28 & -0.39/-0.03 & -0.21/+0.79\td & -0.47\td/-0.38 & -0.35/-0.35 & -0.42\td/-0.01   \\
fo & 76.04/68.63 & +0.19/-0.40 & +0.02/-0.57\td & -0.45/-0.62\td & +0.44/+0.05 & +1.57\td/+0.91\td & -0.33/-0.46   \\
gsw & 52.49/39.27 & +1.45/+0.97 & -1.80/-0.90 & +0.69/+0.42 & +0.21/+0.62 & +0.55/+0.07 & -0.83/-1.25   \\
gun & 18.44/7.81 & +0.61/0.00 & +2.66\td/+1.21 & +3.03\td/+0.68 & +0.23/-0.08 & +2.35\td/+1.14 & +0.23/-0.46   \\
hsb & 60.62/51.99 & +1.24\td/+0.83\td & +1.40\td/+1.41\td & +1.32\td/+1.46\td & +1.31\td/+1.15\td & -0.01/+0.02 & +0.30/+0.18   \\
kk$^*$ & 73.31/61.48 & -0.43/-1.22\td & -0.92\td/-1.12\td & -0.38/-0.83\td & +0.13/-0.23 & +0.14/-0.77\td & -0.39/-1.45\td   \\
kmr & 27.09/12.27 & -0.23/-0.77\td & -1.66\td/-0.97\td & +0.27/-0.12 & -0.78\td/-1.21\td & -1.33\td/-1.80\td & -0.50/-1.20\td   \\
koi & 34.34/20.05 & +0.50/+1.25 & +1.50/+1.25 & -0.75/-1.00 & +3.26\td/+2.76 & +0.25/+1.25 & +1.75/+1.25   \\
kpv & 29.66/15.59 & +0.90/+0.23 & +0.81/+0.26 & -1.24\td/-0.38 & +0.12/+0.17 & -0.61/-0.09 & +0.35/-0.09   \\
krl & 62.18/48.58 & -2.49\td/-2.13\td & -1.81\td/-1.65\td & -1.87\td/-1.81\td & -1.26\td/-1.29\td & -0.55/-0.48 & -1.03\td/-1.23\td   \\
mdf & 35.47/22.99 & 0.00/0.00 & +0.18/+0.18 & +2.14/+1.43 & 0.00/+0.18 & +1.60/+1.43 & -0.71/-0.36   \\
mr$^*$ & 59.95/44.66 & +1.46/+0.97 & -2.43/-3.16\td & -2.67/-2.67 & -0.97/-1.21 & -1.21/-0.97 & -1.21/-2.43   \\
myv & 33.07/18.38 & -1.82\td/-0.61\td & -2.32\td/-0.90\td & -0.47/+0.29 & -0.68\td/-0.12 & -0.84\td/-0.02 & -0.76\td/-0.20   \\
olo & 55.98/41.67 & +0.60/+0.07 & +1.01/+0.27 & -1.08/-1.34 & +0.87/+1.08 & +1.14/+0.81 & +1.14/+0.67   \\
pcm & 46.95/33.93 & -0.54/+0.05 & +1.50\td/+2.65\td & +1.28\td/+1.51\td & -0.44/+0.32 & +0.94\td/+1.87\td & -0.26/+0.09   \\
sa & 39.88/19.05 & -3.85\td/-2.60\td & -3.91\td/-0.98 & -1.63/-0.33 & -2.06\td/-1.14 & -2.22\td/-1.03 & -3.04\td/-1.47\td   \\
ta$^*$ & 65.36/41.68 & +0.05/+0.55 & +2.82\td/+3.27\td & +0.30/+1.21 & +2.71\td/+2.46\td & +2.01\td/+1.51\td & +2.97\td/+2.82\td   \\
te$^*$ & 81.00/67.82 & +0.97/+0.14 & +0.69/+1.11 & 0.00/+0.42 & +0.14/+0.28 & 0.00/-0.14 & -0.69/-0.28   \\
tl$^*$ & 73.97/57.88 & +8.22\td/+9.93\td & +8.90\td/+8.56\td & +5.82\td/+4.45\td & +6.51\td/+7.88\td & +3.77/+4.11\td & +9.25\td/+9.93\td   \\
wbp & 21.02/9.87 & -3.82/-1.59 & -7.64\td/-4.46\td & -0.64/+0.32 & -2.87/0.00 & +0.32/-1.59 & -4.14\td/-2.23   \\
yo$^*$ & 57.28/39.75 & +0.23/+0.26 & +1.20\td/+0.34 & +1.16/+0.98 & -1.99\td/-1.46\td & -0.71/+0.19 & -1.05/-0.79   \\
yue & 50.61/34.21 & +1.03\td/+1.11\td & -0.03/-0.21 & +0.75\td/+0.37 & -0.32/-0.63\td & +0.68\td/+0.86\td & +0.52\td/+0.50\td   \\
 
\hline
    \end{tabular}}
    \caption{Comparison of zero-shot dependency parsing in UAS/LAS between vanilla and bootstrapped mBERT on 28 target languages.
    \textdagger\ indicates statistically significant results from McNemar's test ($p < 0.05$).
    Language codes with asterisks (*) are languages included in the mBERT pre-training data.
    }
    \label{table:parsing-result.all}
\end{table*}

\end{document}